\pdfoutput=1

\documentclass[11pt]{article}

\usepackage[]{ACL2023}

\usepackage{times}
\usepackage{latexsym}

\usepackage[T1]{fontenc}

\usepackage[utf8]{inputenc}

\usepackage{microtype}

\usepackage{inconsolata}

\usepackage{algorithm2e}
\usepackage{makecell}
\usepackage{graphicx} 
\usepackage[export]{adjustbox}
\usepackage{microtype}
\usepackage{placeins}
\usepackage{booktabs}
\usepackage{amssymb}

\title{Do not Mask Randomly: \\
Effective Domain-adaptive Pre-training by Masking In-domain Keywords}

\author{Shahriar Golchin$^\dagger$, Mihai Surdeanu$^\dagger$, Nazgol Tavabi$^\diamond$, Ata Kiapour$^\diamond$ \\
  $^\dagger$University of Arizona, Tucson, AZ, USA \\
  $^\diamond$Harvard Medical School, Boston, MA, USA \\
  \texttt{golchin@arizona.edu} \\\\}

\begin{document}
\maketitle
\begin{abstract}
We propose a novel task-agnostic in-domain pre-training method that sits between generic pre-training and fine-tuning. Our approach selectively masks {\em in-domain keywords}, i.e., words that provide a compact representation of the target domain.
We identify such keywords using KeyBERT \citep{grootendorst2020keybert}.
We evaluate our approach using six different settings: three datasets combined with two distinct pre-trained language models (PLMs).
Our results reveal that the fine-tuned PLMs adapted using our in-domain pre-training strategy outperform PLMs that used in-domain pre-training with random masking as well as those that followed the common pre-train-then-fine-tune paradigm.
Further, the overhead of identifying in-domain keywords is reasonable, e.g., 7--15\% of the pre-training time (for two epochs) for BERT Large \citep{devlin-etal-2019-bert}.\footnote{The code for all of our experiments is available at \url{https://github.com/shahriargolchin/do-not-mask-randomly}.}
\end{abstract}

\section{Introduction}
Employing large pre-trained language models (PLMs) is currently a common practice for most natural language processing (NLP) tasks \citep{tunstall2022natural}.
A two-stage pre-train-then-fine-tune framework is usually used to adapt/fine-tune PLMs to downstream tasks \citep{devlin-etal-2019-bert}.
However, motivated by ULMFiT \citep{howard2018universal} and ELMo \citep{peters-etal-2018-deep}, \citet{Gururangan2020DontSP} showed that incorporating in-domain pre-training (also known as domain-adaptive pre-training) between generic pre-training and fine-tuning stages can lead to further performance improvements in downstream tasks because it ``pulls'' the PLM towards the target domain.
At this intermediate stage, the domain adaptation for PLMs is typically handled by continuing pre-training in the same way, i.e., using randomly-masked tokens on unstructured in-domain data \cite{devlin-etal-2019-bert}. 
Here, we argue that this intermediate pre-training should be performed differently, i.e., masking should focus on {\em words that are representative of target domain} to streamline the adaptation process. 

We propose a novel task-independent in-domain pre-training approach for adapting PLMs that increases domain fit by focusing on {\em keywords} in the target domain, where keywords are defined as ``a sequence of one or more words that offers a compact representation of a document's content'' \citep{rose2010automatic}.
By applying token masking only to in-domain keywords, the meaningful information in the target domain is more directly captured by the PLM.
This is in contrast to the classic pre-training strategy that randomly masks tokens \citep{devlin-etal-2019-bert}, which may overlook domain-meaningful information, or the in-domain pre-training methods that selectively mask tokens deemed important given the downstream task \cite[inter alia]{Gu2020TrainNE}, which require incorporating information from the downstream task into the pre-training stage.
We empirically show that our method offers a better transmission of high-quality information from the target domain into PLMs, yielding better generalizability for the downstream tasks.

The key contributions of this paper are:
{\flushleft \textbf{(1)}} We propose the first task-agnostic selective masking technique for domain adaptation of PLMs that relies solely on in-domain keywords.
In particular, we first extract contextually-relevant keywords from each available document in the target domain using KeyBERT~\cite{grootendorst2020keybert} and keep the most frequently occurring keywords to be masked during the adaptation phase.


{\flushleft \textbf{(2)}} We evaluate our proposed strategy by measuring the performance of fine-tuned PLMs in six different settings.
We leverage three different datasets for text classification from multiple domains: IMDB movie reviews \cite{maas-EtAl:2011:ACL-HLT2011}, Amazon pet product reviews from Kaggle,\footnote{\url{https://www.kaggle.com/datasets/kashnitsky/exploring-transfer-learning-for-nlp}} and PUBHEALTH \citep{kotonya-toni-2020-explainable}.
Our experiments show that the classifiers trained on top of two PLMs---in our case, Bidirectional Encoder Representations from Transformers (BERT) Base and Large \citep{vaswani2017attention,devlin-etal-2019-bert}---that are adapted based on our suggested approach outperform all baselines, including the fine-tuned BERT with no in-domain adaptation, and fine-tuned BERT adapted by random masking.
Further, the overhead of identifying in-domain keywords is reasonable, e.g., 7--15\% of the pre-training time (for two epochs of data) for BERT Large.

\section{Related Work}

Bidirectional Encoder Representations from Transformers (BERT) brought pre-training to transformer networks \cite{vaswani2017attention} through masked language modeling (MLM) \cite{devlin-etal-2019-bert}.
They showed that a simple two-step paradigm of generic pre-training followed by fine-tuning to the target domain can significantly improve performance on a variety of tasks.

However, after showing that infusing an intermediate pre-training stage (commonly known as in-domain pre-training) can help pre-trained Long Short-Term Memory models learn domain-specific patterns better \citep{howard2018universal,peters-etal-2018-deep}, \citet{Gururangan2020DontSP} found that the same advantage applies to PLMs as well.
Since then, several efforts proposed different domain-adaptive pre-training strategies.

 Unsurprisingly, one of the most extensively utilized in-domain pre-training methodologies has been to employ classic random masking to adapt PLMs into several domains \citep{lee2020biobert, beltagy2019scibert, alsentzer2019publicly, tavabi2022natural, tavabi2022building, araci2019finbert}.
Following this, \citet{zheng2020improving} introduced the fully-explored MLM in which random masking is applied to specific non-overlapping segments of the input sequence.
The limitation of random masking that we aim to address is that it may put unnecessary focus on tokens that are not representative of the target domain.

In contrast, task-specific selective masking methods mask tokens that are important to the downstream task.
For each task, ``importance`` is defined differently: \citet{Gu2020TrainNE} let an additional neural model learns important tokens given the task at hand; \citet{ziyadi2020example} defined importance by masking entities for the named entity recognition task, and \citet{feng2018pathologies} found important tokens by input reduction---maintaining model's confidence in the original prediction by reducing input---and they were left with a few (potentially nonsensical) tokens that were treated as important to model.
Similarly, \citet{Li2020TaskspecificOO} designed a task-dependent objective for dialogue adaptation, and \citet{ke2019sentilare} proposed label-aware MLM for a sentiment analysis task.
In the same vein, token selection in certain domains, e.g., biomedical and clinical domains, was performed based on the entities relevant to the domain \cite{Lin2021EntityBERTEM, zhang2020conceptualized, pergola2021boosting}.

Note that other MLM-based pre-training strategies focused on training a language model from scratch \citep[][inter alia]{zhang2020pegasus, joshi2020spanbert, sun2019ernie}. However, since our work focuses on in-domain pre-training, we skip this part for brevity.

In this study, we propose an information-based domain-adaptive pre-training that, without being aware of the downstream task, selectively masks words that are information-dense with respect to the target domain.
As a result, PLMs adapted using our mechanism outperform baselines adapted with random masking or fine-tuned directly.
In the following sections, we refer to our approach as ``keyword masking pre-training.``

\section{Approach}



\subsection{Extracting In-domain Keywords}
In order to extract keywords relevant to the domain of interest, we use KeyBERT \citep{grootendorst2020keybert}.
In a nutshell, KeyBERT uses BERT's \cite{devlin-etal-2019-bert} contextualized embeddings to find the $n$-grams--in our scenario, unigrams--that concisely describe a given document.
In particular, word embeddings with the highest cosine similarity to the overall document-level representation are identified as keywords that best represent the entire document.
We configure KeyBERT to extract up to 10 keywords from each input document.
Note that we do not pre-train or fine-tune BERT as the underlying model for KeyBERT.

\subsection{Removing Noisy Keywords}
\label{sec:removing-noisy}
After extracting domain-specific keywords, we compute the frequency of each specific word that has been recognized as a keyword in all in-domain documents.
Subsequently, we sort them in descending order of their frequency and keep only the most frequent ones. This simple strategy allows us to remove keywords that are likely to be noisy or irrelevant to the target domain.

Figure \ref{fig:1} summarizes the noisy keyword removal process for PUBHEALTH dataset (see Appendix \ref{appendix:b} for other domains).
Note that the actual figure has a very long tail on the right, indicating that the actual in-domain keywords (or parts where information is condensed in the target domain) are frequently repeated.
The graph displays the frequency of terms along with the number of times they are identified as keywords.
In the PUBHEALTH dataset, for example, more than 10,000 words were detected as keywords only once.
Thus, we select the cut-off point where the curve is intended to leap up, signaling that keywords with repetition below the threshold were excluded from the list of domain-relevant keywords.\footnote{The threshold is adjusted via three points: an empirically chosen point from the graph, a point before, and a point after it. Following keyword masking based on each of these three thresholds, we choose the one that resulted in the highest F1 score on the validation split as the final threshold.} 
Namely, in the PUBHEALTH dataset, all words detected fewer than eight times as a keyword were removed from the list of in-domain keywords, and consequently, for performing keyword masking pre-training.
The provided examples on the graph in Figure \ref{fig:1} indicate a qualitative indication that KeyBERT, coupled with our frequency-based heuristic, selects meaningful domain-specific keywords.
For example, our approach identifies relevant keywords (e.g., {\em health}, {\em coronavirus}), while skipping other less relevant ones (e.g., {\em gym}, {\em gift}).


\begin{figure*}[!t]
\centering
\includegraphics[scale=0.32]{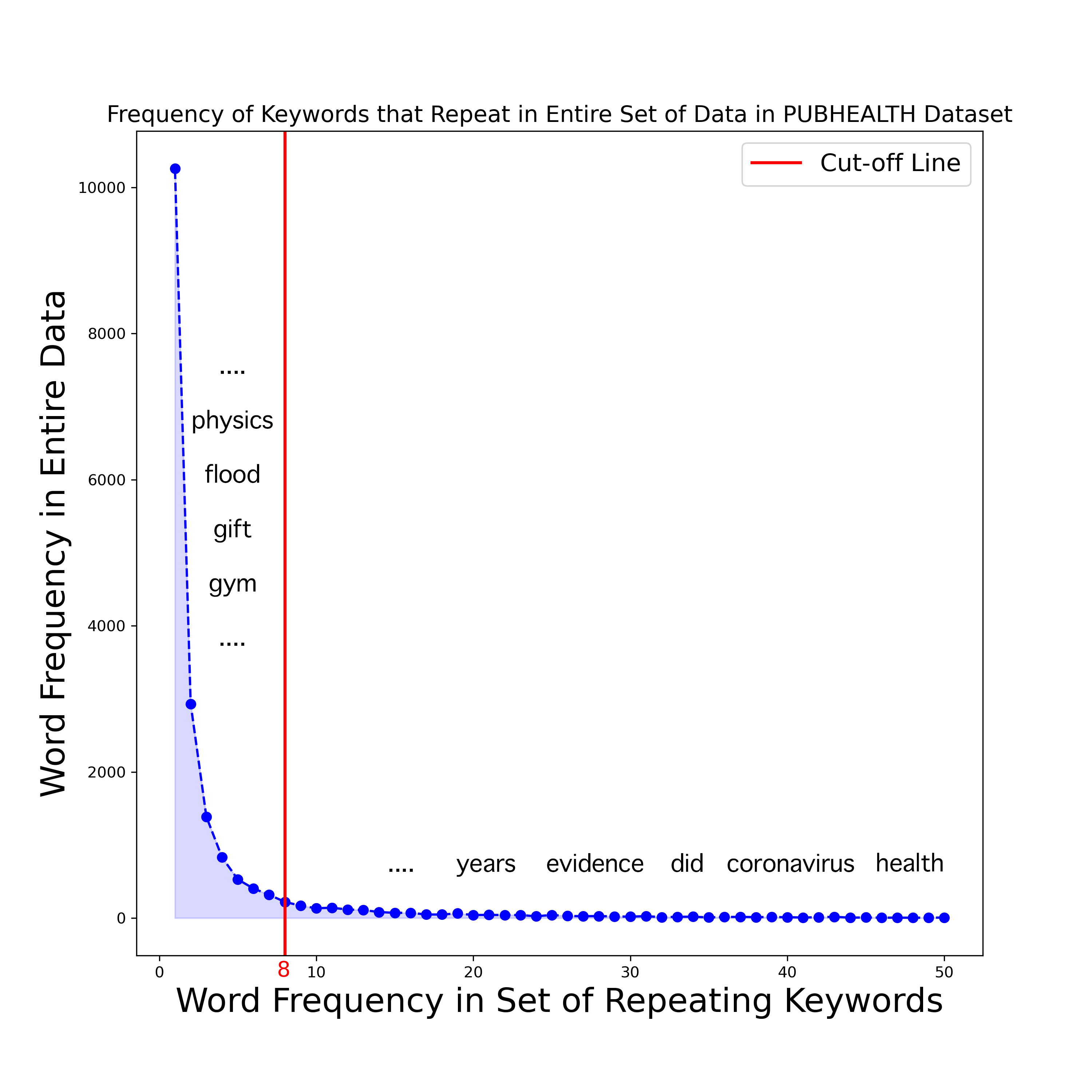}

\vspace{-5mm}
\caption{The graph shows the frequency of the last 50 most frequent keywords in the PUBHEALTH domain and the cut-off line for removing noisy keywords.}
\vspace{-5.8mm}
\label{fig:1}
\end{figure*}


\subsection{Keyword Masking Pre-training}
We pair the list of retrieved candidate keywords with all target domain documents to perform keyword masking pre-training.
If any of the keywords from the list appear in the input documents, the tokens corresponding to those keywords get masked given the masking probability.
In our pre-training strategy, we use a constant learning rate scheduler with a high masking probability rather than a linear one to force the majority of tokens associated with keywords to be masked while continuously learning from surrounding tokens.
As our approach inherits from MLM \citep{devlin-etal-2019-bert}, the tokens related to keywords are masked 80\% of the time, replaced 10\% of the time with other tokens, and left unchanged 10\% of the time.
Note that during pre-training masking only applies to the tokens that match the candidate keywords.
Therefore, there is no pre-training for unmasked tokens.

\subsection{Fine-tuning and Baselines}
We compare the performance of all fine-tuned PLMs adapted using our technique with two other baselines: fine-tuned PLMs adapted using random masking, and fine-tuned PLMs with no in-domain adaptation.
For all these settings we employ both BERT Base and BERT Large \cite{devlin-etal-2019-bert}.

\section{Experimental Setup}
\textbf{Data:} In our experiments, we chose tasks and datasets with sufficient amounts of unlabeled data for the domain adaptation stage in order to observe the effects of keyword selection.\footnote{For example, we did not use the GLUE dataset~\cite{wang2018glue} because the included texts are short.}
In particular, we evaluate our method on three text classification datasets:
PUBHEALTH \cite{kotonya-toni-2020-explainable}, which contains public health claims associated with veracity labels, IMDB movie reviews dataset \cite{maas-EtAl:2011:ACL-HLT2011}, and Amazon pet product reviews dataset (from a Kaggle competition).\footnote{\url{https://www.kaggle.com/datasets/kashnitsky/exploring-transfer-learning-for-nlp}}

Based on the thresholds we studied for filtering out the noisy keywords (see Section \ref{sec:removing-noisy}), we gathered 2,116, 7,274, and 6,881 domain-specific keywords from the PUBHEALTH dataset, IMDB dataset, and Amazon dataset, respectively.

{\flushleft \textbf{Settings:}} 
We use KeyBERT \cite{grootendorst2020keybert} to extract up to 10 unigram keywords per input document utilizing contextualized word embeddings of BERT Base \cite{devlin-etal-2019-bert}, stratified by the Maximal Marginal Relevance (MMR) \cite{mmr} with a threshold of 0.8.

To perform keyword masking pre-training, we set the masking probability to 0.75 with a constant learning scheduler.
The other hyperparameters are left at their default values from the Hugging Face data collator for whole word masking \cite{wolf-etal-2020-transformers}.
For random masking pre-training, we set the masking probability to 0.15, which is a standard value for continual MLM pre-training, and left the remaining hyperparameters at the values provided by the Hugging Face data collator for language modeling \cite{wolf-etal-2020-transformers}. Note that the default learning rate scheduler is linear.
Further, in all settings, pre-training is limited to two epochs, and the batch size of 16 is adopted during both the adaptation and fine-tuning stages.

With the learning rate set to 2e-5 and the weight decay set to 0.01  \cite{devlin-etal-2019-bert}, we fine-tune the whole network for all of our adapted models and baselines for up to four epochs in all datasets, while keeping the other hyperparameters at the default value of Hugging Face \cite{wolf-etal-2020-transformers}.
The models that obtained the highest F1 score in the validation partition are then chosen and evaluated on the test split of the datasets.


\begin{table}[t]
\centering
\begin{footnotesize}
\begin{tabular}{lcc}
\toprule
 \multicolumn{3}{c}{\emph{PUBHEALTH Dataset}} \\ \midrule
 \textbf{Adaptation Method}  & \textbf{Accuracy (\%)} & \textbf{F1 Score (\%)} \\ 
 \midrule
No Adaptation & 64.80 & 63.23  \\
 Random Masking   & 65.77 & 64.94  \\
Our Keyword Masking & \textbf{*66.09} & \textbf{*65.40} \\
\end{tabular}
 \end{footnotesize}
\vspace{-4mm}
\end{table}

\vspace{2.3mm}

\begin{table}[!t]
\centering
\begin{footnotesize}
\begin{tabular}{lcc}
\toprule
\multicolumn{3}{c}{\emph{IMDB Movie Reviews Dataset}} \\ \midrule
 \textbf{Adaptation Method} &  \textbf{Accuracy (\%)} & \textbf{F1 Score (\%)} \\ \midrule
No Adaptation & 94.44 & 94.43 \\ 
Random Masking & 94.96 & 94.95 \\ 
Our Keyword Masking & \textbf{*95.36} & \textbf{*95.35} \\ 
\end{tabular}
 \end{footnotesize}
\vspace{-4mm}
\end{table}

\vspace{0.5mm}

\begin{table}[!t]
\centering
\begin{footnotesize}
\begin{tabular}{lcc}
\toprule
\multicolumn{3}{c}{\emph{Amazon Pet Product Reviews Dataset}} \\ \midrule
 \textbf{Adaptation Method} &  \textbf{Accuracy (\%)} & \textbf{F1 Score (\%)} \\ \midrule
No Adaptation & 85.89 & 85.73  \\ 
Random Masking & 86.33 & 86.31 \\ 
Our Keyword Masking & \textbf{*87.14} & \textbf{*86.98} \\
 \bottomrule
\end{tabular}
 \end{footnotesize}
 \vspace{-2mm}
 \caption{A comparison between the performance of fine-tuning adapted PLMs using our keyword masking and other baselines when {\em BERT Base} is used as the PLM.
 The best results are shown \textbf{bold} and the obtained statistically significant results compared to random masking are indicated by an asterisk  (\textbf{*}) (see Appendix \ref{appendix:f}).}
\vspace{-5.8mm}
\label{table:tbl1}
\end{table}


\begin{table}[t]
\centering
\begin{footnotesize}
\begin{tabular}{lcc}
\toprule
 \multicolumn{3}{c}{\emph{PUBHEALTH Dataset}} \\ \midrule
 \textbf{Adaptation Method} & \textbf{Accuracy (\%)} & \textbf{F1 Score (\%)} \\ 
 \midrule
No Adaptation & 66.42 & \textbf{65.08} \\ 
Random Masking & 63.90 & 64.74 \\ 
Our Keyword Masking & \textbf{*66.66} & 64.74  \\
\end{tabular}
 \end{footnotesize}
\vspace{-4mm}
\end{table}

\begin{table}[!ht]
\centering
\begin{footnotesize}
\begin{tabular}{lcc}
\toprule
 \multicolumn{3}{c}{\emph{IMDB Movie Reviews Dataset}} \\ \midrule
 \textbf{Adaptation Method} & \textbf{Accuracy (\%)} & \textbf{F1 Score (\%)} \\ \midrule
No Adaptation & 95.38 & 95.37  \\ 
Random Masking & 95.50 & 95.49  \\
Our Keyword Masking & \textbf{95.52} & \textbf{95.51}  \\
\end{tabular}
 \end{footnotesize}
\vspace{-4mm}
\end{table}

\begin{table}[!ht]
\centering
\begin{footnotesize}
\begin{tabular}{lcc}
\toprule
 \multicolumn{3}{c}{\emph{Amazon Pet Product Reviews Dataset}} \\ \midrule
 \textbf{Adaptation Method} & \textbf{Accuracy (\%)} & \textbf{F1 Score (\%)} \\ 
 \midrule
No Adaptation & 85.69 & 85.71  \\ 
Random Masking & 86.84 & 86.72 \\ 
Our Keyword Masking & \textbf{*87.58} & \textbf{*87.51} \\
 \bottomrule
\end{tabular}
 \end{footnotesize}
  \vspace{-2mm}
\caption{A comparison between the performance of fine-tuning adapted PLMs using our keyword masking and other baselines when {\em BERT Large} is used as the PLM.
 The best results are shown \textbf{bold} and the obtained statistically significant results compared to random masking are indicated by an asterisk  (\textbf{*}) (see Appendix \ref{appendix:f}).}
\vspace{-5.8mm}
\label{table:tbl4}
\end{table}

\section{Results and Discussion}

Table \ref{table:tbl1} and \ref{table:tbl4} report the performance of fine-tuned models that used multiple domain-adaptive pre-training methods for each of our settings: three different datasets and two distinct PLMs.
Table \ref{table:tbl1} contains the results for {\em BERT Base} as underlying PLM; Table \ref{table:tbl4} uses {\em BERT Large}.

In particular, each table contrasts the performance of two fine-tuned baselines—one without adaptation/in-domain pre-training and one with random masking in-domain pre-training—to a fine-tuned model adapted using our keyword masking.


Both tables show that our approach outperforms all other baselines in all six settings.
The improvements are statistically significant in four out of six settings (Appendix \ref{appendix:f}). 
This highlights the importance of selecting information-carrying keywords for masking during the in-domain pre-training.

The results reveal that our suggested in-domain pre-training technique outperforms alternative settings with or without standard in-domain pre-training on target domain unlabeled data.
Although the benefits of continual pre-training vary depending on the domain and the task at hand \citep{Gururangan2020DontSP}, our adaptation strategy always has a greater impact on PLMs in capturing domain-specific patterns compared to typical random masking when in-domain adaptation has a positive impact on downstream tasks.
This indicates that our pre-training method indeed exposes the PLMs to relevant in-domain representations.

Given the superior outcomes seen in our six different experiments, we can argue that our selective masking strategy, which is task-agnostic as random masking yet more effective, could potentially widely replace random masking in the intermediate pre-training stage for a variety of NLP tasks.
Other than performance, our method is simple and has no ``pathological behavior`` \citep{feng2018pathologies} (see Appendix \ref{appendix:c}).
Additionally, our method takes 2 to 10 minutes of computational overhead to extract keywords. This accounts for 7\% to 39\% of pre-training time of only two epochs (Appendix \ref{appendix:a}).

\section{Conclusion}
We proposed the first task-agnostic selective masking pre-training approach, dubbed ``keyword masking,`` to adapt PLMs to the target domains.
For keyword masking, we first extract in-domain keywords from the target domain using KeyBERT \citep{grootendorst2020keybert}, and after excluding the noisy ones, we only mask the selected keywords during adaptation.

We evaluated our methodology using six different settings.
The results revealed that when in-domain pre-training is conducted using our approach, all fine-tuned PLMs outperform those with no adaptation or adapted using random masking.
Further, we observed that our pre-training approach was superior for difficult tasks, i.e., datasets with many labels and more complexity.
Lastly, keyword masking pre-training can be widely substituted with random masking during shift domain in NLP tasks since it is task-independent, as simple to use as random masking, and more effective.

\section{Limitations}
Although all pre-training approaches require a sufficient amount of data, given how we defined keywords, longer sequences suit our approach better than short ones for studying the effects of keyword selection.
Further, as shown in this study, our findings strongly imply that the strategy we suggested for adapting PLMs can effectively enhance their performance on text classification as the downstream task. To determine whether these findings can translate to other NLP applications, however, further experiments are required.

\section{Ethics Statement}
Although keyword extraction may amplify bias depending on the input documents and the way it extracts keywords, KeyBERT \citep{grootendorst2020keybert} has not been reported to exhibit this behavior. Further work may be necessary to thoroughly explore the potential of introducing undesired bias. 

\bibliography{anthology,custom}
\bibliographystyle{acl_natbib}

\appendix

\section{Cost of In-domain Keyword Extraction}
\label{appendix:a}

For all of our settings, we leverage a single NVIDIA RTX A6000 GPU.
Depending on the size of the datasets and the length of input documents, running KeyBERT \citep{grootendorst2020keybert} for extracting in-domain keywords adds additional computation that can be different between 2 and 10 minutes in our settings.
The overhead time for keyword extraction and the in-domain pre-training time for each of the settings are compared in Table \ref{table:tbl2}.
As can be noticed, the time ratio for keyword extraction to pre-training time ranges from 7\% to 15\% in settings using BERT Large, and 19\% to 39\% for settings with BERT Base, which is reasonable.
Note that when pre-training is performed for more epochs, this ratio noticeably decreases.
The reported ratios are based on only two epochs of in-domain pre-training in our settings.


\begin{table*}[t]
\label{tab:my-table}
\centering
\begin{footnotesize}
\begin{tabular}{lccccc}
\toprule
 \textbf{Dataset Name} &  \textbf{\makecell[c]{BERT Base \\ Pre-train Time}} & \textbf{\makecell[c]{BERT Large \\ Pre-train Time}} &  \textbf{\makecell[c]{Keyword  \\ Extraction Time}} &  \textbf{\makecell[c]{Time Ratio to \\ BERT Base (\%)}} &  \textbf{\makecell[c]{Time Ratio to \\ BERT Large (\%)}} \vspace{-1mm} \\ \midrule
 PUBHEALTH & 4.35 & 11.22 & 1.71 & 39 & 15\\
 IMDB & 29.98 & 79.97 & 9.14 & 30 & 11\\
 Amazon & 38.21 & 100.01 & 7.47 & 19 & 7\\
 \bottomrule
\end{tabular}
\end{footnotesize}
\vspace{-2mm}
\caption{The pre-training time for two epochs, and inference time for KeyBERT in minutes.}
\vspace{-5mm}
\label{table:tbl2}
\end{table*}

\section{Removing Noisy Keywords (Graphs)}
\label{appendix:b}
\begin{figure*}[!htb]
\centering
      \includegraphics[scale=0.31]{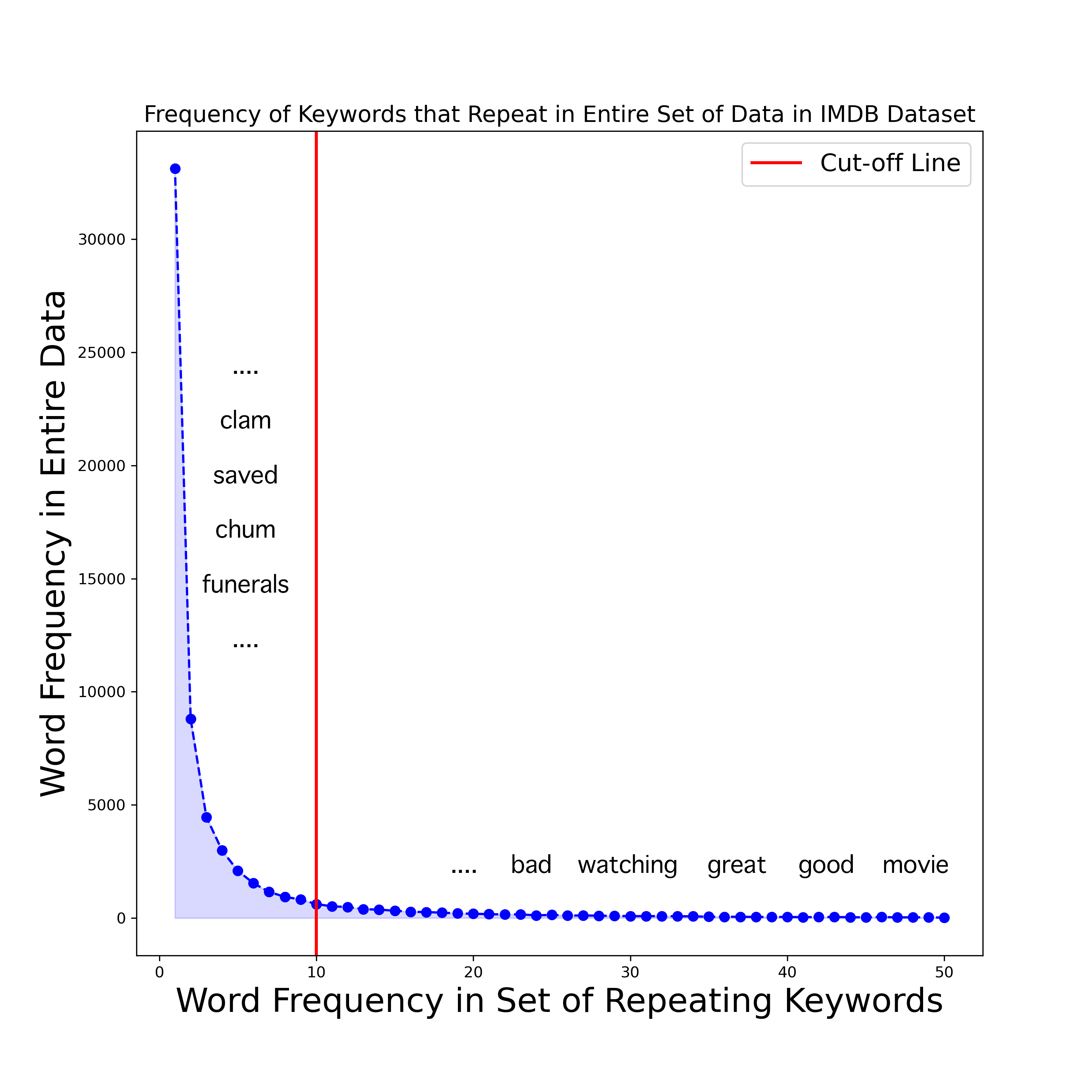}
    \hfill
      \includegraphics[scale=0.31]{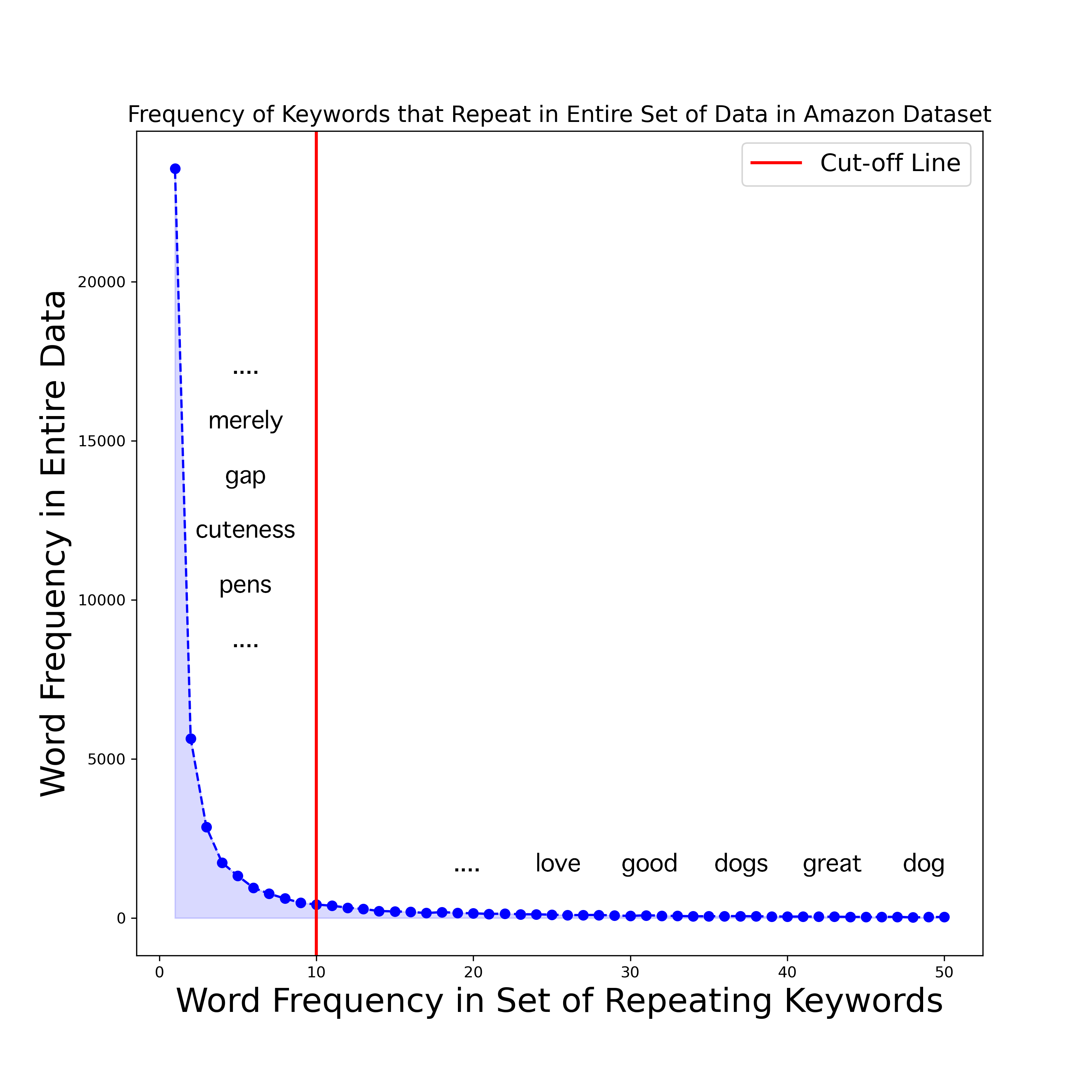}
    \quad
\vspace{-5mm}
\caption{The graphs show the frequency of the last 50 most frequent keywords in IMDB and Amazon datasets along with the cut-off line for removing noisy keywords.
For keyword masking, keywords are selected from the subset to the right of the cut-off line (due to space constraints, we do not show the actual lengthy right tail of the charts).
A few examples of words that were and were not selected as in-domain keywords given this heuristic are shown on the graphs as well.}
\label{fig:2}
\end{figure*}

Similar to Figure \ref{fig:1}, which illustrates the removal of noisy keywords for the PUBHEALTH dataset, Figure \ref{fig:2} displays this procedure for the IMDB dataset and the Amazon dataset.

\section{Pathological-free Behavior}
\label{appendix:c}

It is possible that tokens to be selected for masking are not associated with the domain according to human experts, but they nevertheless yield better downstream classifiers.
For instance, \citet{feng2018pathologies} demonstrated that even when the model is left with a small number of tokens after input reduction, it can still be confident in its predictions even though the left tokens are meaningless.
A similar phenomenon was reported for prompting~\cite{shin2020autoprompt}. To show our masking method's non-pathological behavior, we asked two human annotators to annotate the domain relevance of 50 randomly-chosen words that were selected for masking by the respective method.\footnote{The annotators were two of the authors. The annotations were independent, i.e., no annotator saw the decisions of the other. The names of the methods used to generate the 50 words to annotate were hidden during annotation.} The annotations were performed using a three-point Likert scale: irrelevant, moderately relevant, and relevant. 

Table~\ref{table:tbl3} reports the results of this experiment as well as the Kappa inter-annotator agreement score~\cite{galton1892finger,Cohen1960ACO,10.2307/2531300,McHugh2012InterraterRT}. We draw two observations from this table. First, the agreement between the two annotators is high---substantial or near perfect---which indicates that this task is well-defined. Second, the annotators agreed that the number of moderately- or fully-relevant words is much higher in the keyword-based strategy than in the random masking method. This result further highlights that our masking strategy is indeed relying on the identification of domain-relevant keywords to mask rather than picking up artifacts of the entanglement present in neural architectures \cite{sculley2015hidden}.


\begin{table*}[h!]
\label{tab:my-table}
\centering
\begin{footnotesize}
\newcolumntype{P}[1]{>{\hspace{0pt}}p{#1}}
\begin{tabular}{@{}lcccccccccccc@{}}
\toprule
 \textbf{\makecell[l]{Dataset Name / Masking Method}} & \textbf{\makecell[c]{No. Irrelevant \\ Words}} &  \textbf{\makecell[c]{No.  Moderately \\ Related Words}} &  \textbf{\makecell[c]{No. Related \\ Words}} & \textbf{\makecell[c]{Kappa \\ Value}} &  \textbf{\makecell[c]{Level of \\ Agreement}} \vspace{-1mm} \\
 \midrule
PUBHEALTH / Random Masking & 32 & 6 & 8  & 0.84 & Near Perfect \\ 
PUBHEALTH / Keyword Masking & 14 & 10 & 16  & 0.70 & Substantial \\ 
Amazon / Random Masking & 40 & 5 & 3 & 0.87  & Near Perfect \\ 
Amazon / Keyword Masking & 11 & 17 & 14  & 0.72 & Substantial \\ 
IMDB / Random Masking & 42 & 0 & 4  & 0.65 & Substantial  \\ 
IMDB / Keyword Masking & 11 & 7 & 24  & 0.73 & Substantial \\ 
 \bottomrule
\end{tabular}
 \end{footnotesize}
  \vspace{-2mm}
\caption{The results of measuring inter-rater reliability using Cohen's kappa coefficient for 50 randomly selected words/tokens for masking during in-domain pre-training.}
\vspace{-5mm}
\label{table:tbl3}
\end{table*}


\section{Implementation of Keyword Masking}
To implement our keyword masking strategy, we develop a new data collator by subclassing the Hugging Face data collator for whole word masking \cite{wolf-etal-2020-transformers}.
Our data collator masks only the tokens according to a certain list of keywords given a probability of masking.
Note that our data collator inherits from MLM \citep{devlin-etal-2019-bert}, and no other words or tokens are masked during pre-training except keywords provided by the list.

\section{Statistical Analysis}  
\label{appendix:f}

We analyze the statistical significance of the obtained improvements using a bootstrap resampling technique with 1,000 samples in the resampling process \cite{10.1214/aos/1176344552,EfroTibs93, 10.1214/ss/1063994968}.
The hypothesis that we investigate is if the results achieved by keyword masking are better than the random masking pre-training strategy. We implement two variants of this hypothesis: one compares F1 scores, and the other compares accuracies.


\begin{table}[t!]
\label{tab:my-table}
\centering
\begin{footnotesize}
\newcolumntype{P}[1]{>{\hspace{0pt}}p{#1}}
\begin{tabular}{@{}lcc@{}}
\toprule
 \textbf{Dataset Name / PLM Name} &  \textbf{\makecell[c]{F1 Score \\ $p$-value}} & \textbf{\makecell[c]{Accuracy \\ $p$-value}} \vspace{-1mm} \\ \midrule
 PUBHEALTH / BERT Base & 0.015 & 0.018\\
 PUBHEALTH / BERT Large & 0.505 & 0.010 \\
 IMDB / BERT Base & 0.046 & 0.050 \\
 IMDB / BERT Large & 0.468 & 0.454 \\
 Amazon / BERT Base & 0.000 & 0.000 \\
 Amazon / BERT Large & 0.002 & 0.002 \\ \bottomrule
\end{tabular}
\end{footnotesize}
\vspace{-2mm}
\caption{The computed $p$-values for F1 score and accuracy for each of our settings using bootstrap resampling with 1,000 samples.}
\vspace{-5mm}
\label{table:tbl5}
\end{table}

Table \ref{table:tbl5} lists the results of this analysis.
Overall, the table exhibits that in situations when PLM benefits well from in-domain pre-training, the difference between keyword masking and random masking is statistically significant for both F1 and accuracy scores with $p$-values $\leq 0.05$.
The differences are not statistically significant in the two scenarios: the IMDB dataset with BERT Large and PUBHEALTH dataset with BERT Large (only for F1 score).
This validates our findings since when BERT Large was employed, the results from keyword masking and random masking for the IMDB dataset are quite similar and close to the fine-tuned vanilla PLM.
Similarly, in the PUBHEALTH dataset with BERT Large, keyword masking and random masking tie in the F1 score, making the difference in the F1 score not statistically significant.
These results further confirm that the benefits of in-domain pre-training vary depending on the domain and the task at hand \cite{Gururangan2020DontSP}; however, when in-domain pre-training has a positive impact on performance and causes significant improvement compared to non-adapted setting, our approach outperforms random masking and yields statistically significant gains.

\section{Detailed Description of Datasets}
\noindent \textbf{PUBHEALTH Dataset} \, The PUBHEALTH dataset is divided into three sections: train, test, and validation.
Samples in each partition are public health claims with one of four veracity labels including false, unproven, true, or mixture.
The labels were assigned by domain experts based on an explanation that they provided for every claim, available in a separate column.
These explanations serve as in-domain unstructured data for our use.
9,832 samples in the train split, 1,225 samples in the validation split, and 1,235 samples in the test split form our dataset after a few unlabeled samples were removed.\footnote{This dataset contains a small number of claims that did not fall under any of the four aforementioned veracity labels.} 

\medskip

\noindent \textbf{IMDB Movie Reviews Dataset} \, The two portions of the IMDB dataset are labeled and unlabeled reviews, each having 50,000 reviews.
The train, validation, and test splits are generated by dividing the labeled portion by 80\%, 10\%, and 10\%, respectively. That is,
40,000 reviews are allotted to the train split and 5,000 each to the validation and test splits. The unlabeled 50,000 reviews are used for pre-training.

\medskip

\noindent \textbf{Amazon Pet Product Reviews Dataset} \, There are six different labels for reviews in the Amazon pet product dataset used in the Kaggle competition: dogs, fish aquatic pets, cats, birds, bunny rabbit central, and small animals.
The dataset contains four splits: train, test, validation, and unlabeled.
However, since the test split does not include labels, we create our own test split by randomly choosing a portion of the train split that is equal in size to the validation split.
As a result, in our setting the validation and test splits each includes 17,353 samples; the train split contains 34,704 samples.
In addition, there are 100,000 reviews without labels in the dataset's unlabeled portion that serve as pre-training data.

\label{sec:appendix}

\end{document}